\documentclass{article}

\usepackage{arxiv}

\usepackage[utf8]{inputenc} %
\usepackage[T1]{fontenc}    %
\usepackage{hyperref}       %
\usepackage{url}            %
\usepackage{booktabs}       %
\usepackage{amsfonts}       %
\usepackage{nicefrac}       %
\usepackage{microtype}      %
\usepackage{lipsum}
\usepackage{listings}
\usepackage{amsmath}
\usepackage{graphicx}
\usepackage{subfigure}
\usepackage{enumitem}
\usepackage{authblk}
\usepackage[numbers]{natbib}
\usepackage{float}
\usepackage[format=plain,
            labelfont={bf,it},
            textfont=it]{caption}

\graphicspath{{Images/}}

\usepackage{listings}
\usepackage{xcolor}
\usepackage{soul}
\definecolor{codegreen}{rgb}{0,0.6,0}
\definecolor{codegray}{rgb}{0.5,0.5,0.5}
\definecolor{codepurple}{rgb}{0.58,0,0.82}
\definecolor{backcolour}{rgb}{0.95,0.95,0.92}

\lstdefinestyle{mystyle}{
    backgroundcolor=\color{backcolour},   
    commentstyle=\color{codegreen},
    keywordstyle=\color{magenta},
    numberstyle=\tiny\color{codegray},
    stringstyle=\color{codepurple},
    basicstyle=\ttfamily\footnotesize,
    breakatwhitespace=false,         
    breaklines=true,                 
    captionpos=b,                    
    keepspaces=true,                 
    numbers=left,                    
    numbersep=5pt,                  
    showspaces=false,                
    showstringspaces=false,
    showtabs=false,                  
    tabsize=2
}

\lstdefinelanguage{csv}{}
\lstdefinelanguage{text}{}

\title{\textbf{Adapting Pretrained Language Models For Solving Tabular Prediction Problems In The Electronic Health Record}}

\author[1,2,4,5]{Christopher McMaster}
\author[1,2,3]{David FL Liew}
\author[4,5]{Douglas EV Pires}

\affil[1]{Department of Clinical Pharmacology \& Therapeutics, Austin Health, Melbourne, Australia}
\affil[2]{Department of Rheumatology, Austin Health, Melbourne, Australia}
\affil[3]{Department of Medicine, University of Melbourne, Melbourne, Australia}
\affil[4]{The Centre for Digital Transformation of Health, University of Melbourne, Melbourne, Australia}
\affil[5]{School of Computing and Information Systems, University of Melbourne, Melbourne, Australia}

\begin{document}
\maketitle
\begin{abstract} \normalsize
  We propose an approach for adapting the DeBERTa model for electronic health record (EHR) tasks using domain adaptation. 
  We pretrain a small DeBERTa model on a dataset consisting of MIMIC-III discharge summaries, clinical notes, radiology reports, and PubMed abstracts. 
  We compare this model's performance with a DeBERTa model pre-trained on clinical texts from our institutional EHR (MeDeBERTa) and an XGBoost model. 
  We evaluate performance on three benchmark tasks for emergency department outcomes using the MIMIC-IV-ED dataset. 
  We preprocess the data to convert it into text format and generate four versions of the original datasets to compare data processing and data inclusion. 
  The results show that our proposed approach outperforms the alternative models on two of three tasks (p<0.001) and matches performance on the third task, with the use of descriptive columns improving performance over the original column names.
\end{abstract}


\vfill

\section{Introduction}
Tabular data is ubiquitous in real world data science problems. 
Whilst a large amount of data is naturally generated and collected in tables, such as internet traffic metadata, numerical data from scientific experiments and nursing observations in the intensive care unit, the historically poor performance of many machine learning algorithms on unstructured data has motivated the development of data wrangling methods to convert even non-tabular data into tabular format for optimal algorithm performance. 
But data wrangling and cleaning to convert unstructrued data into tabular data can result in the loss of information, particularly where the data is entered as free text. 
In healthcare applications, this often involves encoding drugs, pathology results and diagnoses as factors in data tables, even when these data are entered in free text fields. 
For example, when developing a predictive model using medication prescription data, the process of data wrangling might involve encoding the medications as factors, or categorising them into groups, such as antihypertensives, anticoagulants, and antiplatelets.
Some details might be retained, such as dose and/or frequency of administration, but often the full rich semantic information present in the free text is lost, such as the route of administration, the indication, tapering or titration schedules, and a detailed history of previous medications for each indication.

Pretrained language models have demonstrated excellent performance on a variety of tasks, including text classification, question answering, and natural language inference\cite{devlin2018bert,liu2019roberta,he2021deberta}.
These models have been trained on large amounts of text data, and have been shown to be able to learn rich representations of language, including the semantic meaning of words and phrases.
Recent work has shown that these models can be used to solve tabular prediction problems by converting tabular data into strings of tokens \cite{Dinh2022-jb,Hegselmann2022-je,ptab}.
However, there are known limitations with how pretrained language models are used for numeric tokens.
\citet{wallace-2019} demonstrated that language models trained to decode token embeddings of numbers (i.e. predict the floating point number from its embedding) can only accurately recapture within the range of numbers they were trained on.
This could introduce significant error when applied to numeric data outside of the training range.

In the setting of electronic health record (EHR) prediction problems where the data is a 20:80 mix of tabular and unstructured (including free text)\cite{Kong2019-nn}, we investigate whether language models can be used to solve tabular prediction problems without the need for extensive data wrangling.
Given data privacy concerns limiting cloud service use and limited local compute, we test whether competitive performance can be achieved with small language models.
Because pure tabular datasets of EHR data often involve the transformation of free text into categorical factors, we also investigate whether retaining the original free text data improves performance.
We also investigate strategies for handling numeric data in the tokeniation and embedding process.
Finally, given the abundance of free text notes in the EHR, we evaluate performance with and without further pretraining on free text EHR data.

\section{Related work}

Several authors have proposed using \textbf{language models for tabular data}. \citet{ptab} proposed a method for using BERT \cite{devlin2018bert} to solve tabular prediction problems, by converting tabular data into strings of tokens, performing masked language modelling (MLM) on the resulting strings and then fine-tuning on the target task.
The method was shown to be competitive with state-of-the-art tree-based methods on a variety of pure tabular prediction problems.
LIFT \cite{Dinh2022-jb} and TabLLM \cite{Hegselmann2022-je} use large pretrained large language models to tackle tabular data problems, with a focus on few-shot performance. 
These methods have demonstrated not only competitive overall performance with best-in-class tree-based methods, but also excellent few-shot performance. 
LIFT uses GPT-J\cite{mesh-transformer-jax} (6 billion parameters) and GPT-3\cite{Brown2020-ki} (off-site finetuning via OpenAI API), while TabLLM uses T0pp\cite{Sanh2021-ld} (11 billion parameters). The size of GPT-J and T0pp, and the inavailability of GPT-3 weights, limit their usability in settings with low compute, particularly where data privacy concerns present a barrier to train and deploy models on a cloud service.

\citet{levin2022} benchmarked two different \textbf{tabular transformer architectures}\cite{Huang2020-xc,Gorishniy2021-oy} on multiple EHR tasks from the MetaMIMIC repository\cite{Grzyb2021-mp}.
The authors explored a number of pretraining regimes and found that supervised pretraining on multiple diagnostic tasks improves downstream performance on a different diagnostic task.
Whilst these models do use transformer architectures, they are not language models and therefore cannot be used for tasks with mixed tabular and free text data.
There has been a growing interest in using deep learning methods for working with tabular data in the context of clinical research. Previous studies have shown that convolutional neural networks (CNNs) can be effective for automatically extracting features from tabular data, and have demonstrated improved performance on a variety of tasks, such as predicting patient outcomes and identifying disease subtypes\cite{wang2019tabular,chen2020graph}.

\section{Methodology}

\subsection{Models}

Our proposed approach is based on the DeBERTa model\cite{he2021deberta,he2021debertav3}, a variant of BERT\cite{devlin2018bert} that has been shown to perform well on tasks related to natural language processing in the biomedical domain\cite{McMaster2022-dj}. 
DeBERTa is a deep learning model that uses a transformer architecture\cite{vaswani2017attention} to learn contextual representations of text, allowing it to capture the rich semantic information present in clinical data.
As a baseline model, we used a pretrained DeBERTa small model (version 3)\footnote{https://huggingface.co/microsoft/deberta-v3-small} (\textbf{DeBERTa Small}), a 44 million parameter model.
We then pretrained this model for domain adaptation on the masked language modelling task, using a dataset consisting of 700k MIMIC-III discharge summaries, clinical notes and radiology reports\cite{Johnson2020-vs,Johnson2016-jm}, plus 1 million abstracts from the PubMed dataset\cite{pubmed} published by the National Library of Medicine (\textbf{Deberta Small [adapted]}). 
This allowed the model to learn a wide range of clinical and biomedical concepts with the goal of improving performance on downstream EHR tasks. 
Additionally, we compared performance with a DeBERTa base model (verison 1) that we have previously pretrained on 1.3 million clinical texts from our institutional EHR at Austin Health\cite{McMaster2022-dj} (\textbf{MeDeBERTa}). 
For comparison, we trained \textbf{XGBoost}\cite{chen2016xgboost} models as strong baselines, given the consistently good performance of XGBoost on tabular data tasks. 
We also include the results from the best performing model for each task in the original paper.

\subsection{Data Processing}

To use DeBERTa for working with tabular data, we first preprocess the data to convert it into text format. This involves refining the column headings to make them more descriptive and then concatenating these with the values. Where several columns naturally belong to a list (e.g. one-hot-encoded comorbidity columns), these can be combined into a comma-separated list of features. For example, the following comma-separated values (CSV) file,

\begin{lstlisting}[language=csv]
cci_DM2,cci_Cancer1,age
1,1,50
0,1,73
\end{lstlisting}

would be converted into the following text representations:

\begin{lstlisting}[language=text]
Age in years: 50, Comorbidities: Diabetes Mellitus with Complications, Cancer Without Metastases
Age in years: 73, Comorbidities: Cancer Without Metastases
\end{lstlisting}

To evaluate the performance of our proposed approach, we used the Medical Information Mart for Intensive Care IV Emergency Department (MIMIC-IV-ED) dataset\cite{Goldberger2000-bn,Johnson2020-wt,Johnson2022-ag} to derive three benchmark tasks for emergency department outcomes, as described by \citet{Xie2022-yx}. These tasks are as follows:
\begin{itemize}[noitemsep]
    \item \textbf{Task 1}: The hospitalization outcome predicts whether a patient will be admitted to the hospital following an ED visit.
    \item \textbf{Task 2}: The critical outcome predicts whether a patient will require urgent care, such as transfer to an ICU within 12 hours.
    \item \textbf{Task 3}: The ED reattendance outcome predicts whether a patient will return to the ED within 72 hours after their previous discharge.
\end{itemize}
To generate the datasets for these tasks, we reuse the code from \citet{Xie2022-yx}\footnote{https://github.com/nliulab/mimic4ed-benchmark}.
In their code, the authors use snake case for column names, using abbreviations and shorthand to encode longer descriptions (e.g. "n\_ed\_90d" for the number of emergency department presentations in the previous 90 days). 
Additionally, the authors used a rule-based method to extract 10 common chief complaints from the free text description (abdominal pain, dizziness, etc.), converting this into 10 one-hot-encoded columns representing the presence or absence of each of these complaints.

We generated 4 versions of the original datasets to compare data processing and data inclusion:
\begin{itemize}[noitemsep]
    \item \textbf{Values only}: Text strings of comma-separated values (e.g. "99.4, 76, 16")
    \item \textbf{Original columns}: Text strings of original column names and values separated by a colon, with name/value pairs joined in a semi-colon separated list (e.g. "triage\_temperature: 99.4; triage\_heartrate: 76; triage\_resprate: 16")
    \item \textbf{Descriptive columns}: Text strings of more descriptive column names and values separated by a colon, with name/value pairs joined in a semi-colon separated list (e.g. "triage temperature: 99.4; triage heart rate: 76; triage respiratory rate: 16")
    \item \textbf{Descriptive columns, text problems}: The same as above, with the free text chief complaint added as an extra data point.
\end{itemize}

The datasets were split into train and test sets, as per the original paper.

\subsection{Task Finetuning}

Models were finetuned on each of the three task separately. Finetuning was performed by splitting the predefined training sets into train (98\%) and validation (2\%) sets. 
Each model was trained for 20 epochs, with evaluation performed and model checkpoints saved after every 500 steps. 
The checkpoint with the best performance on the validation set was then loaded and evaluation performed on the predefined test set defined by \citet{Xie2022-yx}.

The XGBoost models were trained using the same training and test sets. 
Hyperparameters were tuned on the training set using 5-fold cross validation and Bayesian optimisation.

\subsection{Model Comparison}

Models were compared using the area under the receiver operating characteristic curve (AUC) metric, with bootstrapped confidence intervals calculated using the method used in the original paper.
To compare the performance of the different data processing methods, we used the two-sample t-test.

\subsection{Explainability}

To investigate the contributions of each token at the individial level and across samples, we calculated layer integrated gradients with respect to the embedding layer\cite{Sundararajan2017-os, Mudrakarta2018-mh} using the Captum library\cite{kokhlikyan2020captum}. Here, the integrated gradient for the $i$th token is calculated by taking the path integral of the layer gradients between two embeddings $x$ and $x'$, where $x'$ is the embedding of the pad token and $x$ is the embedding of the token of interest:

\begin{equation}
    \label{eq:lig}
    IG_i = (x_i - x'_i) \times (\int_{0}^{1} \frac{\partial \mathbf{F}(x' + \alpha (x - x'))}{\partial x_i} d\alpha)
\end{equation}

In practice, the integral is approximated by taking the average of the gradients at $n$ points along the path between $x$ and $x'$. Averaging these gradients across the embedding dimensions gives a token-specific attribution score.

To investigate variable-level attribuition, we classify each token by the variable to which it belongs (e.g. in "Co morbidities : Co agul opathy , Dementia" the following tokens are comorbidity tokens: "Co", "agul", "opathy", "Dementia"). We then calculate the maximum value of the attribution scores for each token to get a variable-level attribution score. We do this for a random sample of 1,000 samples from the test set, and then average the attribution scores across these samples.

\section{Results}
\label{sec:results}

\subsection{Performance}

In all three tasks, we were able to improve on the performance of the best performing model from the original publication (see Table \ref{table1}). The addition of column names improved performance for every task, although improvements were small for tasks 1 and 2. 
More descriptive column names had a smaller impact on performance, however these column names produced shorter tokenized sequences (owing to the tokenizer vocabulary). 
To minimize sequence length, the descriptive column names were retained when adding the free text chief complaint as an additional data point. 
This addition improved performance across all three tasks, although the improvement was largest for task 1.

\begin{table}[H]
      \centering
        \resizebox{\textwidth}{!}{
            \begin{tabular}{llrrlll}
            \toprule
            & Model & DeBERTa Small & DeBERTa Small [adapted] & MeDeBERTA & XGBoost & Original Paper\cite{Xie2022-yx} \\
            Task & Dataset & & & & \\
            \midrule
            1 & Descriptive columns, text problem & 0.864 (0.862, 0.866) & \textbf{0.865 (0.862, 0.867)} & \textbf{0.866 (0.864, 0.868)	} & - & - \\
            & Descriptive columns & 0.827 (0.825, 0.83)	 & 0.827 (0.824, 0.829) & - & - & - \\
            & Original columns & 0.828 (0.826, 0.831) & 0.827 (0.825, 0.83) & - & - & - \\
            & Values only & 0.828 (0.825, 0.83) & 0.827 (0.825, 0.83) & - & 0.828 (0.828, 0.828) & 0.822 (0.821, 0.825) \\
            \hline
            2 & Descriptive columns, text problem & 0.909 (0.904, 0.913) & \textbf{0.91 (0.906, 0.914)} & \textbf{0.911 (0.907, 0.915)	} & - & - \\
            & Descriptive columns & 0.889 (0.884, 0.893) & 0.889 (0.884, 0.894) & - & - & - \\
            & Original columns & 0.889 (0.884, 0.893) & 0.888 (0.883, 0.892) & - & - & - \\
            & Values only & 0.886 (0.882, 0.891) & 0.888 (0.883, 0.892) & - & 0.891 (0.891, 0.891) & 0.883 (0.879, 0.888) \\
            \hline
            3 & Descriptive columns, text problem & 0.702 (0.690, 0.714) & \textbf{0.705 (0.693, 0.716)} & 0.699 (0.687, 0.711)	 & - & - \\
            & Descriptive columns & 0.689 (0.676, 0.702) & 0.690 (0.677, 0.704) & - & - & - \\
            & Original columns & 0.694 (0.682, 0.707) & 0.694 (0.681, 0.707) & - & - & - \\
            & Values only & 0.653 (0.639, 0.664) & 0.665 (0.651, 0.68) & - & \textbf{0.703 (0.703, 0.703)} & 0.700 (0.691, 0.713) \\
            \bottomrule
            \end{tabular}
        }
      \caption{\label{table1}Area Under the Receiver Operating Characteristic (ROC-AUC) for Each Task}
\end{table}

Pretraining DeBERTa on a combination of MIMIC-III and PubMed resulted in small improvements, seen most prominently in task 3. Overall, our previously published MeDeBERTa model using the free text chief complaint achieved the highest ROC-AUC on tasks 1 and 2 \citet{Xie2022-yx} by 0.043 (P<0.001) and 0.031 (p<0.001) respectively. On task 3, although the mean ROC-AUC was 0.007 higher for the DeBERTa Small [adapted] model, this difference was not statistically significant (p=0.538).

\begin{figure}[h!]
  \centering
  \includegraphics[width=\textwidth]{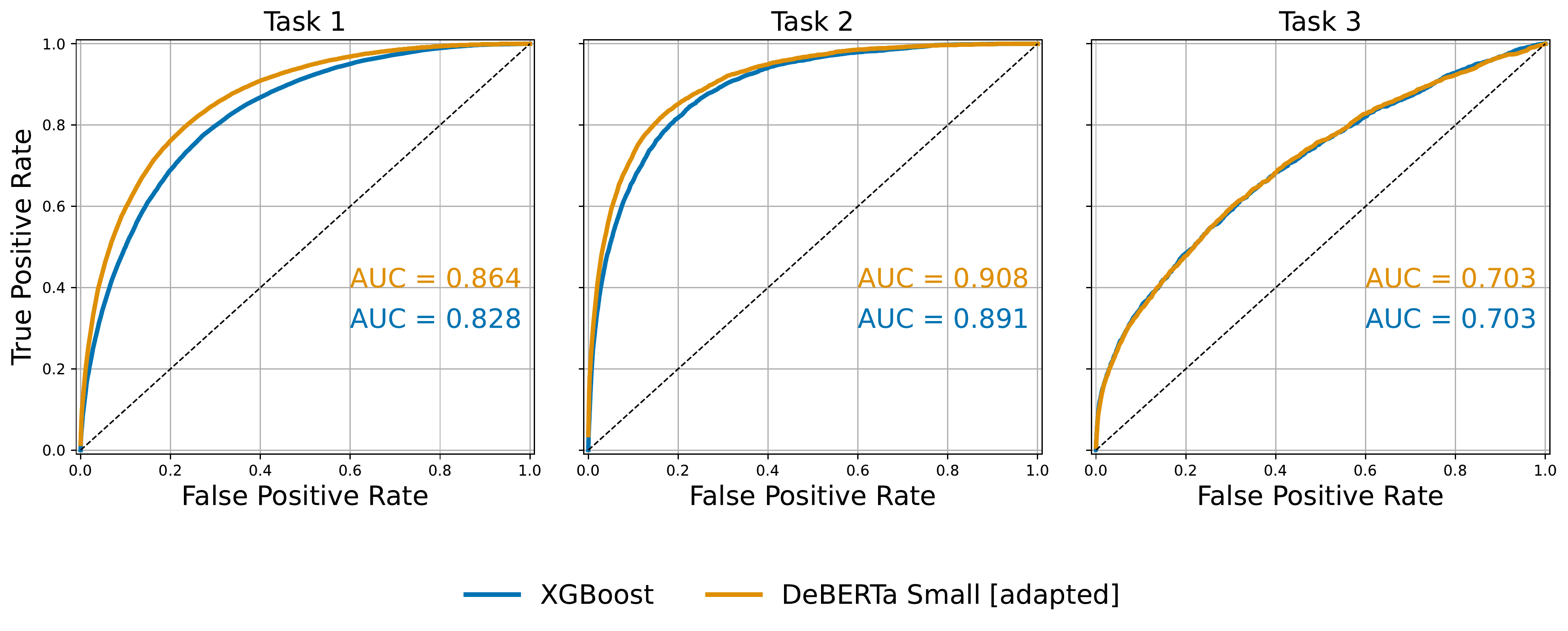}
  \caption{Receiver Operating Characteristic Curves for Each Task}
  \label{fig:roc}
\end{figure}

\subsection{Explainability}

Tokenwise attribution scores were calculated using the embedding gradients to interrogate the contributions of different variables and column headings.
Table \ref{table2} demonstrates a visualization of the attribution scores for 2 different example patients, with and without the free text chief complaint field.
Whilst the model (DeBERTa Small [adapted] in this case) correctly predicted hospitalization outcome for both patients, in the absence of a chief complaint that fit any of the 10 predefined categories, the model attribution scores for the free text chief complaints were high and contributed to higher model confidence.

\begin{table}[H]
  \centering
  \includegraphics[width=\textwidth]{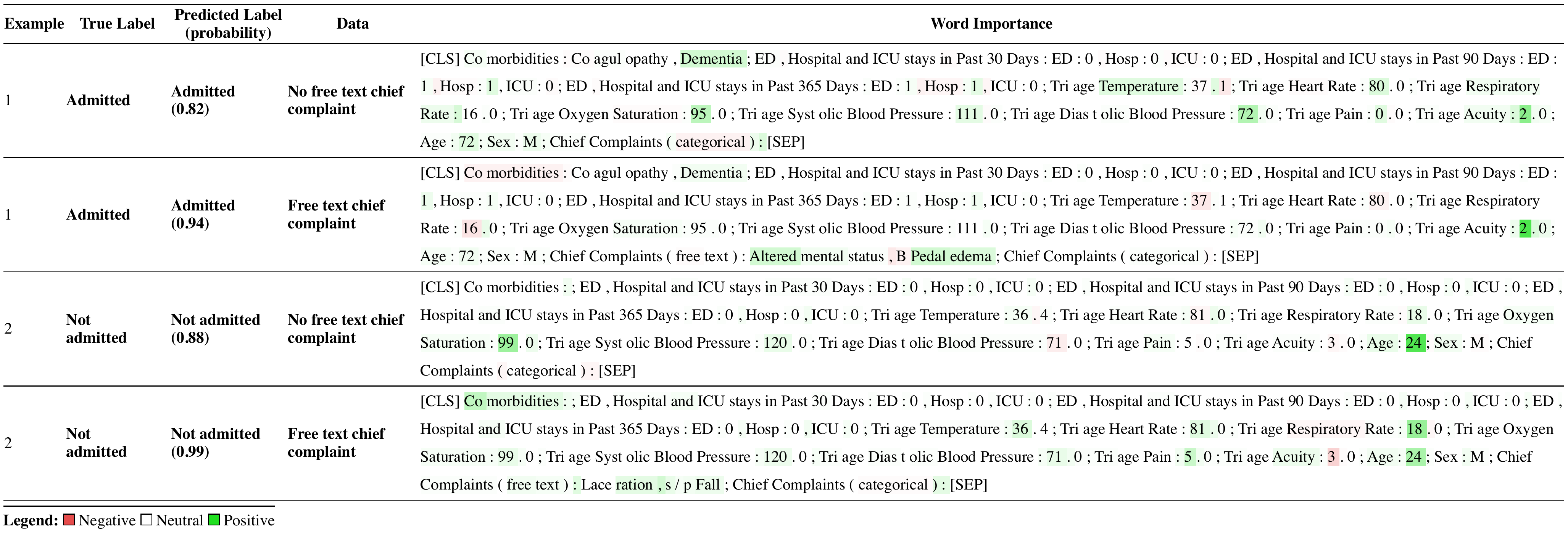}
  \caption{Word importance for two example patients with model predictions for \textbf{task 1} (hospitalization), with and without the free text chief complaint.}
  \label{table2}
\end{table}

\begin{figure}[H]
  \centering
  \includegraphics[width=\textwidth]{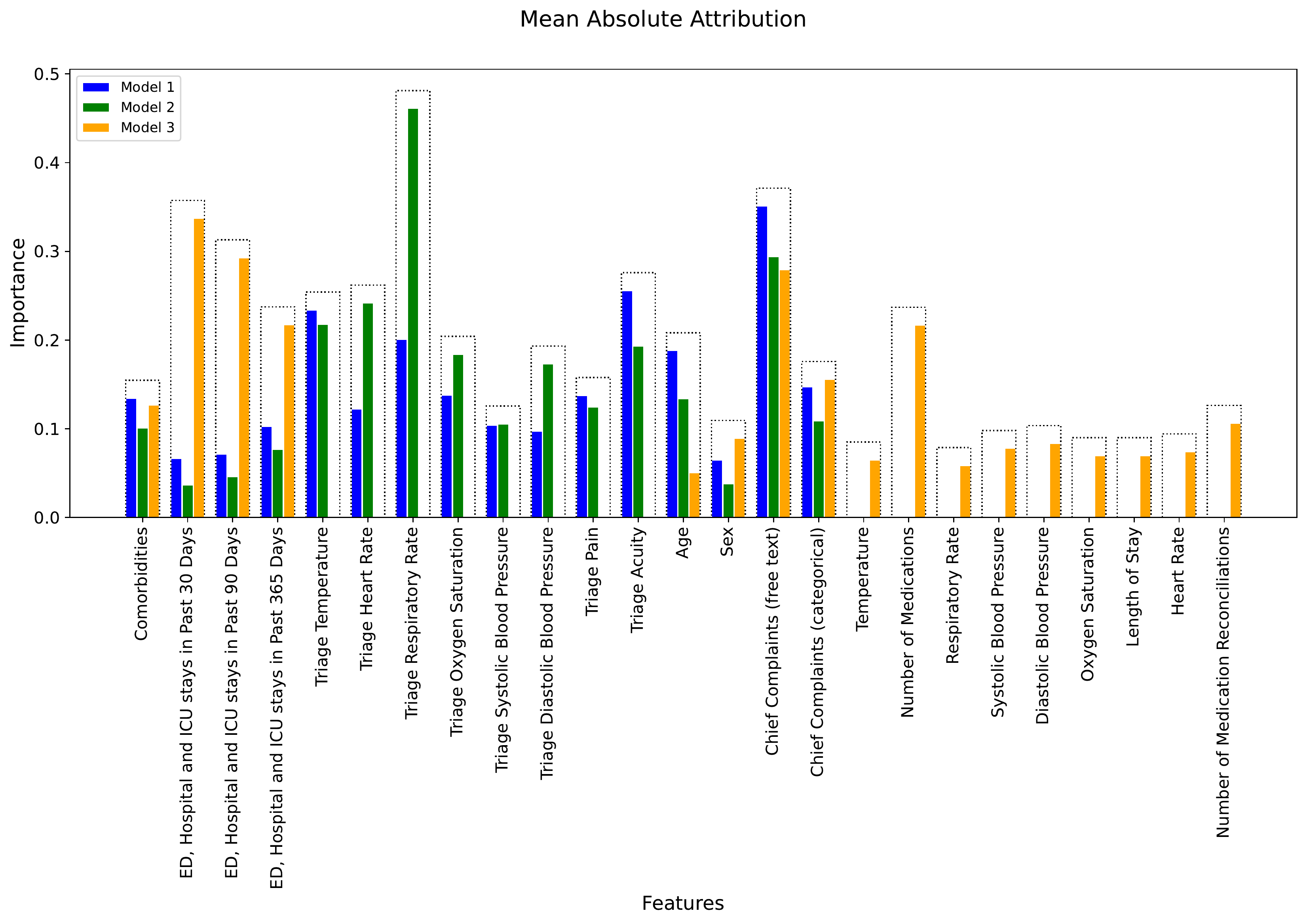}
  \caption{Mean attribution scores across 5,000 randomly sampled patients in the test set for models 1, 2 and 3.}
  \label{fig:att}
\end{figure}

Grouping tokens into variables headings, 5,000 examples from the test set were sampled and the average maximum attribution score was calculated for each variables (Figure \ref{fig:att}). 
The free text chief complaints was, on average, the variable with the largest attribution scores.

\section{Discussion}
\label{sec:discussion}

In this study, we evaluated the effectiveness of our proposed approach, which is based on the DeBERTa model, in predicting emergency department outcomes. 
We found that the DeBERTa small [adapted] model outperformed the baseline DeBERTa small and XGBoost models on all three benchmark tasks and showed similar performance to a larger DeBERTa model pretrained on a different corpus of clinical notes.

Our results showed that the inclusion of free text data and the use of the \"descriptive columns\" version of the dataset resulted in the best model performance. 
This highlights the importance of data processing and the ability to mix free text and tabular data in EHR datasets.
The ability to mix free text and tabular data allows for a more comprehensive representation of contributing risk factors, which can lead to improved model performance and decision-making.

We demonstrated that small DeBERTa models can achieve competitive performance when fine-tuned on EHR datasets, which is important for compute-constrained settings such as hospitals.
For comparison, a 6 billion parameter model like GPT-J\cite{mesh-transformer-jax} will require approximately 22GB of RAM using single-precision floating point numbers for the weights alone, whereas the DeBERTa small model used in this study requires approximately 0.17GB of RAM.

We used attribution scores to determine the importance of different input features in our machine learning models \ref{fig:att}. 
The attribution scores align with our understanding of the factors leading to the three outcomes we predicted. 
For example, the free text chief complaint was the most important variable in predicting the likelihood of hospitalization, which is consistent with the fact that the decision to admit a patient depends largely on the type of presentation. 
The attribution scores for the models predicting the likelihood of needing urgent care showed a reliance on patient vital signs, particularly respiratory rate, which is consistent with the fact that urgent care is often used to support patients with physiological decompensation. 
Finally, predicting the likelihood of representation within 72 hours was strongly attributed to the number of prior ED, hospital and ICU stays, particularly in the prior 30 days, which is consistent with the fact that patients with a history of frequent visits are more likely to have future visits.

Attribution scores, particularly when determined at the token level, may hold significant clinical value in shedding light on the factors influencing patient outcomes and pinpointing potentially modifiable risk factors. 
The explainability provided by these scores may help in guiding the appropriateness and impact of clinical interventions.
For instance, within patient subgroups where EHR notes exhibit high attribution scores related to medications (including dosing, frequency and indication), it is plausible that deprescribing interventions could have a disproportionately positive impact. 
By identifying patients with high polypharmacy or elevated sedative/anticholinergic burdens, healthcare professionals may leverage these attribution scores to optimize medication regimens, reduce adverse drug interactions, and improve overall patient safety.

Although the general approach of using small language models for mixed structured and free text EHR data is promising, we have not presented a comprehensive evaluation on a wide range of tasks and have not directly compared our approach to much larger models.
We are planning to conduct a more comprehensive evaluation of our approach on a wider range of tasks and compare performance to larger models.

We have presented an approach for adapting the DeBERTa model for EHR tasks using domain adaptation. 
Our proposed approach outperforms other models on three benchmark tasks for emergency department outcomes using the MIMIC-IV-ED dataset. 
We also compared our model's performance with a DeBERTa model pre-trained on clinical texts from our institutional EHR and an XGBoost model. 
Our results demonstrate the effectiveness of our approach and highlight the importance of data preprocessing and inclusion. 
Our study provides a foundation for future work in adapting transformer-based models for EHR tasks and highlights the potential of these models for improving prediction models and therefore outcomes for hospitalised patients.

\clearpage

\bibliographystyle{unsrtnat}  
\bibliography{references.bib}

\end{document}